\documentclass[runningheads]{llncs}
\usepackage{graphicx}
\usepackage{enumerate}
\begin{document}
\title{Deep Learning Approach for Intelligent Named Entity Recognition of Cyber Security}

\titlerunning{Deep Learning based Named Entity Recognition}
%
%\titlerunning{Abbreviated paper title}
% If the paper title is too long for the running head, you can set
% an abbreviated paper title here
%

\author{Simran K\inst{1}\and Sriram S\inst{1} \and
Vinayakumar R\inst{2, 1}\and
Soman KP\inst{1}}
\authorrunning{Simran et al.}
% First names are abbreviated in the running head.
% If there are more than two authors, 'et al.' is used.
%
\institute{Center for Computational Engineering and Networking, Amrita School Of Engineering, Amrita
vishwa vidyapeetham, Coimbatore, India. \email{simiketha19@gmail.com, sri27395ram@gmail.com} \and Division of Biomedical Informatics, Cincinnati Children's Hospital Medical Centre, Cincinnati, OH, United States.\\ \email{Vinayakumar.Ravi@cchmc.org, vinayakumarr77@gmail.com}}

\maketitle             % typeset the header of the contribution

\begin{abstract}
In recent years, the amount of Cyber Security data generated in the form of unstructured texts, for example, social media resources, blogs, articles, and so on has exceptionally increased. Named Entity Recognition (NER) is an initial step towards converting this unstructured data into structured data which can be used by a lot of applications. The existing methods on NER for Cyber Security data are based on rules and linguistic characteristics. A Deep Learning (DL) based approach embedded with Conditional Random Fields (CRFs) is proposed in this paper. Several DL architectures are evaluated to find the most optimal architecture. The combination of Bidirectional Gated Recurrent Unit (Bi-GRU), Convolutional Neural Network (CNN), and CRF performed better compared to various other DL frameworks on a publicly available benchmark dataset. This may be due to the reason that the bidirectional structures preserve the features related to the future and previous words in a sequence.

\keywords{Information Extraction \and Named Entity Recognition \and Cyber Security \and Deep learning.}
\end{abstract}
\section{Introduction}

Technology has evolved tremendously in the previous couple of decades making diverse online web sources (news, forums, blogs, online databases, etc.) available for common use. Newfound information regularly seems first on these kinds of online web sources in unstructured text format. Mostly, it requires a ton of time, sometimes years, for this information to properly classify into a structured format. However, cases related to Cyber Security does not have that kind of time, so timely extraction of this information plays an imperative job in numerous applications. Information modeling of cyber-attacks can be used by many individuals  such as auditors, analysts, researchers, etc \cite{1}. At the core, knowledge extraction task is recognition of named entities of a domain, such as products, versions, etc. The best accuracy given by the current Named Entity Recognition (NER) devices, in this area of Cyber Security, depend on feature engineering. However, feature engineering has many issues, for example, depending on look-ups or dictionaries to recognize known entities \cite{35}. With profoundly dynamic fields such as Cyber Security, it is difficult to fabricate as well as harder to keep up the word references and in turn, consumes a lot of time for creating these NER tools. The limitation of these tools is that they are domain-specific and cannot accomplish good accuracy when applied to any other domain. These issues can be overcome by using Part of Speech (POS) tagging. POS tagging is a task of identifying the words as nouns, adjectives, verbs, adverbs, and so on (Distributed Denial of Service (DDoS) is mostly taken as a noun). POS tagging can also diminish the time delay in this domain.

Conditional Random Fields (CRFs) emerged in recent years as the most successful method for entity extraction. CRFs are best for tagging and sequential data as they can predict sequences of labels for a set of input samples, taking the context of the input data into account. Mikolov et al. \cite{3} proposed a method called word2vec which converts every word in a corpus into a low dimensional dense vector representation. These vectors can reflect the semantic relationship between the words which was not possible in classical one-hot vectors. For example, the difference between the vectors representing the words ‘king’ and ‘queen’ is similar to the difference between the vectors representing the words ‘man’ and ‘woman’. These relationships result in the clustering of semantically similar words in the vector space.

Another leap forward in the ongoing years is the Deep Learning (DL) field. DL is an enhanced classical neural network model which naturally learns non-linear combinational features on its own whereas classical methods, for example, CRFs can just learn linear combinations of the defined features. This reduces the user work of tedious feature engineering \cite{1,extra1,extra2,extra3}. The expansion in information accessibility, increase in hardware processing power particularly GPUs, different activation functions, etc has made DL exceptionally practical and feasible. Recent days, the applications of DL have been leverage towards various applications in the field of Cyber Security \cite{41,42,43,44}.

The hybrid model of Long Short Term Memory (LSTM) and CRF architecture proposed by Lampal et al.  \cite{6} for NER is a blend of LSTM, CRFs, and word2vec model. As this architecture is domain and entity type agnostic, it very well can be connected to any area as long as the input is an annotated corpus of the same format as the CoNLL-2000 dataset. This architecture is applied to the domain of Cyber Security NER in  \cite{9}. As the corpus of Cyber Security is not widely available, they utilized the corpus created in the work done by Bridges et al.  \cite{1} for training the model. The author compared the performance of LSTM-CRF with CRFSuite and LSTM-CRF beat CRFSuite with two percent accuracy. 

The major contributions are:

\begin{enumerate}
     \item This work proposes a deep learning based framework for NER in Cyber Security.
     \item Various deep learning architectures are evaluated on well known and benchmark data set.
     \item The importance of non-linear text representation is discussed in the context of Cyber Security NER.
\end{enumerate}

The remaining paper is arranged as follows. Section 2 documents the related works followed by background related to deep learning architectures in Section 3. Section 4 describes the details of the data set used and Section 5 provides a description of proposed architecture. Section 6 reports the experiments, results, and observation whereas section 7 will conclude this paper with remakes on future work of research.

\section{Related Works}

A lot of efforts are made for automatically labeling Cyber Security entities in the past years. Concepts in Common Vulnerability Enumeration (CVE-3) were annotated to instantiate a security ontology in \cite{10}. Reuters \cite{11} likewise did with the assistance of his web journals and OpenCalais. Mulwad et al.  \cite{12} searched the web to identify security relevant text and then train an SVM classifier to identify the descriptions of potential vulnerability. Afterward utilizing OpenCalais alongside the taxonomy of Wikipedia to identify and classify vulnerability entities. In spite of the fact that these sources rely upon standard entity recognition software, they are not prepared to recognize domain specific ideas. This is because of the general nature of their training data  \cite{13,14}. Joshi et al.  \cite{15} noted comparative discoveries where NERD system  \cite{16}, Open-Calais and Stanford Named Entity Recognizer were all commonly helpless to recognize Cyber Security domain entities. This is due to the fact that these tools do not utilize domain-specific corpus. Stanford NER system with hand-labeled domain specific training corpus was created in  \cite{15} and delivered better outcomes for the greater part of their domain-specific entities.

Joshi et al.  \cite{15} additionally performed hand annotating for a small corpus. This small corpus was then encouraged to Stanford NER \textquotedblleft off-the-shelf \textquotedblright \hspace{0.3mm} layout for train a CRF entity extractor  \cite{13}. Their work has recognized the equivalent Cyber Security issue yet they didn't address a more broad issue of automatically the labeling process.

Semi-supervised methods are additionally utilizing entity extraction rather than supervised methods as they function admirably with very small training data also. \cite{23} have used semi-supervised methods for entity extraction and have got beneficial outcomes. A bootstrapping algorithm created by  McNeil et al.  \cite{24} depicts a prototypical execution to extract data about vulnerabilities and exploits.  \cite{25} utilizes known databases for seeding bootstrapping strategies.  \cite{26} developed an exact strategy to automatically label text from different data sources and then provide public access to a data set which is annotated with Cyber Security entities. \cite{27} proposed an algorithm for extracting security entities and their relationships using semi-supervised NLP with a small amount of input data. The precision of 0.82 was achieved using small corpus. Either rule-based or machine learning  \cite{28} approaches are utilized for NER and regularly the two are mixed  \cite{29}. Rule-based is a blend of lookups and rules for pattern matching that is hand-coded by an area expert. These rules use the contextual information of the entity to decide if candidate entities are substantial or not.  Machine learning based NER approaches utilize an assortment of models, for example, Maximum Entropy Models  \cite{30}, Hidden Markov Models (HMMs)  \cite{31}, CRFs  \cite{32}, Perceptrons  \cite{33}, Support Vector Machines (SVMs)  \cite{34}, or neural networks  \cite{35}. The best NER approaches incorporate those dependent on CRFs. CRFs address the NER issue utilizing an arrangement labeling model. In this model, the label of an entity is demonstrated as reliant on the labels of the previous and following entities in a predefined window.

As deep neural networks address many short comes of classical statistical technique such as feature engineering, they are considered as a potential alternative \cite{36}. As feature are learned automatically by the neural network, they decline the measure of human endeavors required in various applications. Significant enhancements in accuracy in neural network feature with respect to hand-engineered features are shown by empirical outcomes over a wide arrangement of domains. Recurrent Neural Networks (RNNs) property of having long time memory as they process input with variable length results in outstanding accomplishment with several NLP tasks  \cite{37}. In \cite{38}, LSTM property to permit the learning between arbitrary long-distance dependencies enhances its performance over RNN.

In  \cite{39}, LSTM-CRF technique was applied to the issue of NER in the Cyber Security domain utilizing the corpus made accessible by Joshi et al. \cite{15}. Given an annotated corpus of a decent size, it was observed that this technique outflanks the state of the art methods given an annotated corpus of a decent size.

\section{Background}
\subsection{Convolutional Neural Network (CNN)}

CNN is very famous network as it uses less learnable parameters. This network consists of three types of layers which are convolution layers, pooling layers, and fully connected layers. In the convolution layer of this neural network, convolution operation is performed using several filters which slide through the input and learns the features of the input data. The pooling layer is utilized to decrease the size of the previous layer. Min pooling, max pooling, and average pooling are the different types of pooling layers available. The network is connected with a dense layer at the end so that the features can be mapped to the classification.

\subsection{ Recurrent Neural Network (RNN)}
RNN is a very important deep learning architecture which can handle sequential information whereas feedforward neural networks fail at doing this. That's why RNN is utilized for applications related to sequential data, for example, speech data as it can learn multi-variation like PoS tagging  \cite{40} using the hidden units present in the recurrent cells. The issue of exploding or vanishing gradients is the main problem with RNN.

\subsection{ Long Short-Term Memory (LSTM)}

LSTM was introduced to solve the problem of vanishing gradient in RNN. It selectively remembers the patterns for a long duration of time. The error that will be backpropagated through layers is preserved which is a big help to LSTM. Because of the three gates in LSTM, it has the ability to handle long term as well as short term information whereas RNN can handle only short term data. The cell takes decisions about when to allow read, write, etc. operations as well as what to store and erase.

\subsection{Gated Recurrent Unit (GRU)}

GRU is an extended version of standard RNN and are also considered as a minor variation from LSTM. An LSTM without an output gate is basically called as GRU. It has update gate and reset gate which chooses what data should be passed through them. LSTM can easily perform unbounded counting, so in this way, it is stronger than GRU. That's why the LSTM is able to learn simple languages whereas GRU fails at this.

\subsection{Bidirectional architectures}

To understand the context better and resolve ambiguities in the text, bidirectional recurrent structures are utilized which learns the information from the previous and future time stamps. The two types of connections are one going forward in time and the other going backward in time. These connections help in learning the previous and future representations respectively. The module in this bidirectional recurrent structure could be an RNN, LSTM or GRU. 

\section{Description of the Data set}

Auto-labeled Cyber Security domain text corpus provided by Bridges et al. \cite{1} comprising of around 40 entity types was used in this work.  All the corpora is stored in a single JSON file and high-level JSON element are used to represent each corpus. Each word in these corpora is auto-annotated with an entity type. The files were additionally changed over into CoNLL2000 format in order to feed the data into the model. Every word was annotated in a separate line and separation between the three corpora were removed. When the preprocessing is completed, each line in the data set consists of a word as well as its entity type. The data set was divided into three subsets that is training, validation, and testing consisting of 70\%, 10\%, and 20\% data respectively.

\section{Proposed Architecture}

\begin{figure}[!htbp]
  \centering
  \includegraphics[width=8cm,height=7cm]{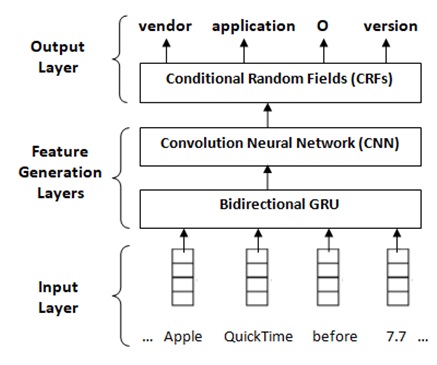}
  \caption{Proposed Architecture.}
  \label{Fig:1}
\end{figure}

The proposed architecture for Cyber Security named entity recognition is shown in Figure \ref{Fig:1}. The architecture has three different sections, namely input, feature generation, and output. In the input layer, the text data is converted into numerical features using Keras embedding. These numerical features are high dimensional in nature so they are passed into a bidirectional GRU layer. Bidirectional GRU layer refines the vectors and feeds them to a CNN layer. CNN network generates more optimal features and feeds it to a CRF layer for enhanced learning. This final layer uses the Viterbi algorithm to generate the most possible tag for the input word.

\section{Experiments, Results, and Observations}

TensorFlow\footnote{https://www.tensorflow.org/} with Keras\footnote{https://keras.io/} were used to implement all the deep learning architectures. All the models are trained on GPU enabled TensorFlow. Various statistical measures are utilized in order to evaluate the performance of the proposed classical machine learning and deep learning models. 

\iffalse

The classification results can be seen in detail using a confusion matrix. Confusion matrix is a matrix representation which shows the quantity of correctly or incorrectly classified data in different classes.

Rows of the matrix represent the instances of actual class whereas columns represent the instances in the predicted class or vice versa. The dimensions of the confusion matrix for binary classification are True Positive, True Negative, False Positive, and False Negative.

True Positive (TP) denotes the quantity of positive samples that are correctly classified.

True Negative (TN) denotes the quantity of negative samples that are correctly classified.

False Positive (FP) denotes the quantity of negative samples that are misclassified.

False Negative (FN) denotes the quantity of positive samples that are misclassified.

Metrics like Precision, Recall, F1-score can be calculated using the confusion matrix.

Precision measure is the ratio of true positive with respect to all the positives predicted.
\begin{equation}
{\rm{Precision}} = \frac{TP}{TP + FP}
\end{equation}

Recall measure is the ratio of true positive with respect to a total number of true actual classifications.
\begin{equation}
{\rm{Recall}} = \frac{TP}{TP + FN}
\end{equation}

F1-score is given by the harmonic mean between precision and recall.
\begin{equation}
{\rm{F1-Score}} = 2 \times \frac{ (Recall \times Precision)}{ (Recall + Precision)}
\end{equation}

\fi

Initially, to identify the optimal architecture for auto-labeled Cyber Security domain text corpus we proposed an architecture which has been explained in Section 5 and we compared the performance of the proposed method with the architectures already published. Various deep learning architectures were implemented and their performance was compared with Bidirectional GRU+CNN-CRF. Bidirectional GRU+CNN-CRF outperformed all other architectures. LSTM is computationally expensive whereas GRU is computationally inexpensive which can lead to the same performance in some cases and some cases the performance of the GRU is better than LSTM. 

The training dataset was used to train the models and these trained models were tested using the testing data set. LSTM, RNN, GRU, Bidirectional GRU contains 128 hidden units. Embedding dimension of 128, learning rate of 0.005, and Stochastic Gradient Descent (SGD) learning method are used in all the experiments. The models were compared in terms of precision, recall, and f1-score and the detailed results are reported in Table \ref{Tab:1}. The performance metrics in terms of precision, recall, and f1-score shows that Bidirectional GRU+CNN-CRF outperformed all other architectures.

\begin{table}[!htbp]
\renewcommand{\arraystretch}{1.3}
\centering
\caption{Average weighted performance metrics for all entity types.}
\label{Tab:1}
\scalebox{0.8}{\begin{tabular}{|l|l|l|l|}
\hline
\textbf{Model}                 & \textbf{Precision} & \textbf{Recall} & \textbf{F1-score} \\ \hline
\textbf{LSTM-CRF} \cite{9}            & 85.3             & 94.1              & 89.5             \\ \hline
\textbf{CRF} \cite{9}           & 82.4             & 83.3              & 82.8                    \\ \hline
\textbf{CNN-CRF}             & 83.1             & 93.9              & 88.2                \\ \hline
\textbf{RNN-CRF}            & 83.5             & 85.6              & 84.5                \\ \hline
\textbf{GRU-CRF}       & 86.5             & 95.7                  & 90.9                \\ \hline
\textbf{Bidirectional GRU-CRF}            & 88.7              & 95.4              & 91.9                \\ \hline
\textbf{Bidirectional GRU+CNN-CRF}  & \textbf{90.8 }            & \textbf{96.2 }              &\textbf{ 93.4}                \\ \hline

\end{tabular}}
\end{table}

\section{Conclusion and Future work}

This work evaluates the efficacy of various deep learning frameworks along with CRF model for NER Cyber Security text data and Bidirectional GRU+CNN-CRF model performed better than other deep learning architecture. The best part about the proposed architecture is that it does not require any feature engineering. This work can be extended for Relation Extraction (RE). RE basically attempts to find occurrences of relations among entities. Product vulnerability description can be better understood by using RE. For instance, RE can be used to identify and differentiate the product that is being attacked and the product that is meant to attack by using the information extracted from the product vulnerability description.

\section*{Acknowledgment}

This research was supported in part by Paramount Computer Systems and Lakhshya Cyber Security Labs. We are grateful to NVIDIA India, for the GPU hardware support to research grant. We are also grateful to Computational Engineering and Networking (CEN) department for encouraging the research.

\end{document}